\documentclass{article}
\usepackage{spconf,amsmath,graphicx}
\usepackage{amssymb}
\usepackage{enumitem}
\setlist{nosep, leftmargin=14pt}

\usepackage{mwe} 
\usepackage{multirow}

\title{FM-Recon: Exploring Foundation Model Features for Accelerated Cardiac Cine MRI Reconstruction}
\name{Donghang Lyu$^{1}$, Marius Staring$^{1}$, Ruochen Gao$^{1}$, Jochen Keupp$^{2}$, Hildo J. Lamb$^{1}$, Mariya Doneva$^{2}$}

\address{$^{1}$Department of Radiology, Leiden University Medical Center, Leiden, The Netherlands\\
$^{2}$Philips Innovative Technologies, Hamburg, Germany
}
\begin{document}
%
\maketitle
\begin{abstract}
Foundation models have demonstrated strong feature extraction capabilities across various medical imaging tasks, yet their potential for MRI reconstruction remains largely unexplored, particularly within unrolled reconstruction networks and under artifact-corrupted inputs. In this work, we explore whether and how features from pretrained foundation models can benefit accelerated cardiac cine MRI reconstruction. To this end, we propose FM-Recon, a framework that refines and integrates extracted features from frozen image encoder of foundation models into an unrolled reconstruction network. Through extensive experiments on the CMRxRecon2023 dataset, we systematically evaluate foundation model features across different reconstruction settings. Our results show that foundation models can extract informative representations from artifact-corrupted inputs and improve reconstruction performance, particularly at moderate acceleration factors. These findings demonstrate the potential of foundation model features for accelerated MRI reconstruction and provide insights into their effective integration into unrolled reconstruction networks.
\end{abstract}
\section{Introduction}
\label{sec:intro}
Cardiac cine magnetic resonance imaging (MRI) is widely used for assessing cardiac function by providing high soft-tissue contrast and dynamic visualization of heart motion. However, long acquisition times and repeated breath-holds can cause patient discomfort and limit imaging efficiency. Accelerated acquisition can alleviate these limitations but introduces undersampling artifacts, making accurate reconstruction essential for preserving image quality.

In recent years, foundation models (FMs)~\cite{1,2,3}, pretrained on large-scale datasets, have demonstrated strong generalization across a wide range of vision tasks. Their success has motivated increasing adoption in medical imaging~\cite{6,21}, where pretrained FM encoders can provide rich representations for downstream tasks~\cite{5}. Such features have been incorporated through different fusion mechanisms, including attention~\cite{10} and matrix-based operations~\cite{11}. Despite this progress, the potential of FM features for MRI reconstruction remains largely unexplored.

In general, integrating foundation models into MRI reconstruction poses two key challenges. First, reconstruction networks differ substantially from conventional vision architectures. Modern approaches commonly employ unrolled networks~\cite{12} due to the iterative nature of traditional reconstruction algorithms. Integrating high-dimensional FM features across cascades can therefore incur considerable computational overhead, especially for dynamic cine MRI. Second, reconstruction begins with zero-filled images corrupted by undersampling artifacts, raising the question of \emph{whether pretrained FMs can extract informative representations from such degraded inputs and how these features can be efficiently refined and integrated to benefit reconstruction.}

To investigate these issues, we conduct an exploratory study on the use of FM features for cardiac cine MRI reconstruction by proposing \textbf{FM-Recon}, a framework for their effective integration into unrolled reconstruction networks. In FM-Recon, features are extracted from a frozen FM image encoder and refined with lightweight adapters~\cite{27} and channel attention~\cite{17}. Then Cross-FiLM modules enable efficient bidirectional interaction between the refined FM representations and native reconstruction features across cascades, enabling feature integration without incurring substantial computational overhead. Furthermore, we investigate different foundation models, cascade depths, and acceleration factors to provide a comprehensive assessment of the effectiveness of FM features under varying reconstruction settings.

Our main contributions are threefold: 
(1) we explore the feasibility of leveraging FM features for cardiac cine MRI reconstruction by introducing FM-Recon, a framework that refines and integrates FM features into an unrolled reconstruction network; (2) we systematically investigate the effects of cascade depth and acceleration factor, characterizing the conditions under which FM features remain informative and improve reconstruction; and (3) we evaluate FM-Recon on the CMRxRecon2023 dataset~\cite{14} using diverse foundation models, demonstrating the general potential of FM feature integration for accelerated cardiac cine MRI reconstruction.

\section{Methods}
\label{sec:format}


\subsection{Problem Formulation}
Given undersampled multi-coil k-space data $\mathbf{y}\in\mathbb{C}^{T\times N_c\times H\times W}$, where $T$ is the number of time frames, $N_{c}$ is the number of coils, $H$ and $W$ denote the height and width of each frame, cardiac cine MRI reconstruction aims to recover $\mathbf{x}\in\mathbb{C}^{T\times H\times W}$ by solving
\begin{equation}
\label{eq1}
    \hat{\mathbf{x}}
    =\operatorname*{argmin}_{\mathbf{x}}
    \|\mathbf{y}-A\mathbf{x}\|_2^2
    +\lambda\mathcal{R}(\mathbf{x}),
\end{equation}
where $A=M\mathcal{F}S$ denotes the forward operator consisting of the undersampling mask $M$, Fourier transform $\mathcal{F}$, and coil sensitivity maps $S$. In deep unrolled reconstruction, $\mathcal{R}$ is typically implemented as a learnable neural network that acts as a regularizer to enforce image priors.
\begin{figure}[tb]
\centering
\includegraphics[width=\linewidth]{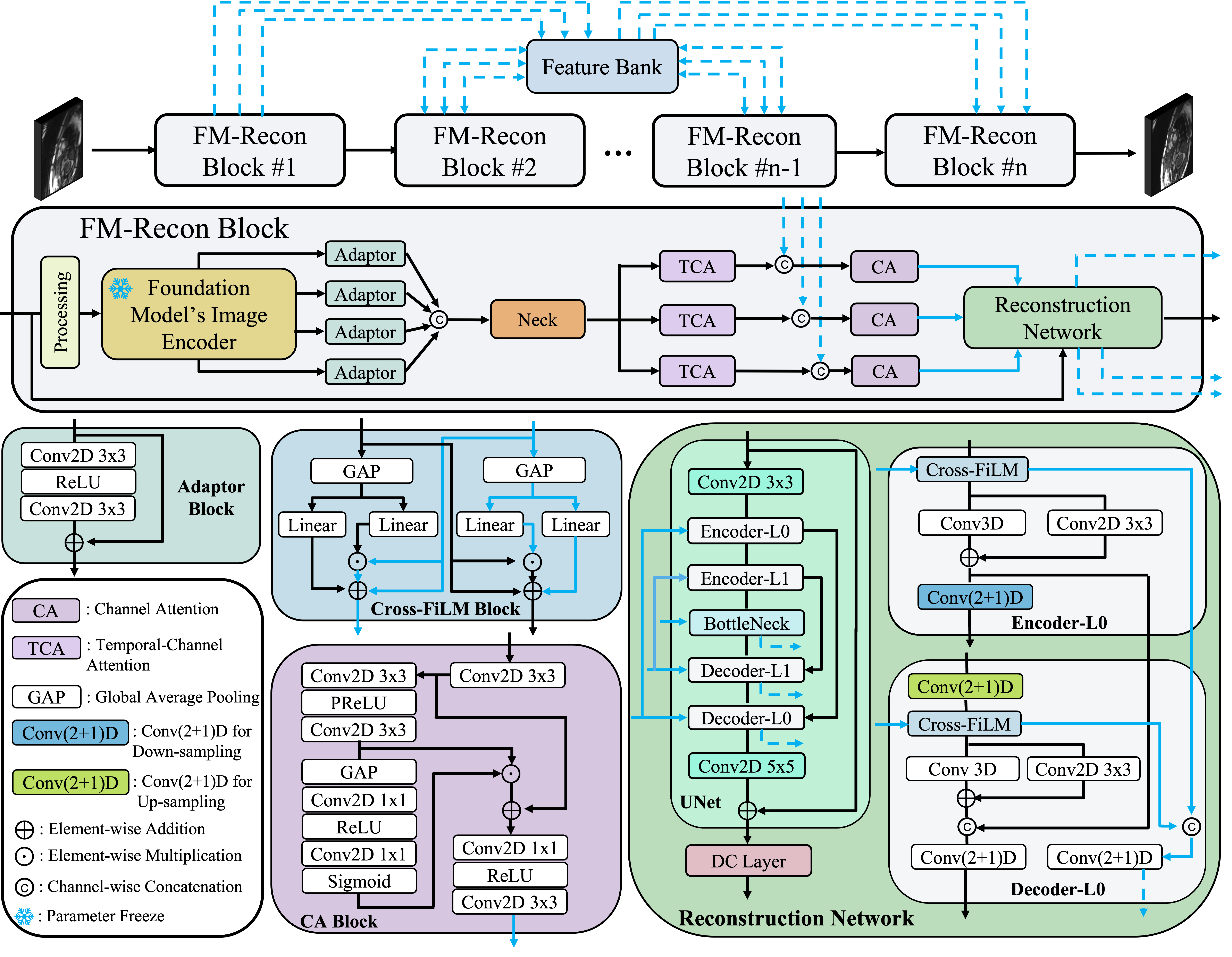}
\caption{Overview of FM-Recon. Frozen FM features are refined and integrated with native reconstruction features through Cross-FiLM modules. Solid blue lines indicate the flow of refined FM features, while dashed blue lines denote bidirectional feature interaction. Updated features are propagated across cascades through the feature bank.}
\label{fig1}
\end{figure}
As illustrated in Fig.~\ref{fig1}, FM-Recon adopts an unrolled architecture and aims to integrate extracted FM features into the reconstruction process. The $i$-th cascade is formulated as
\begin{equation}
    \mathbf{x}_{i+1},f_i =
    \operatorname{FM\text{-}Recon}_{\theta_i}
    (\mathbf{x}_i,\mathbf{y},S,M,\mathcal{B}_i),
\end{equation}
where $\mathbf{x}_i$ is the current reconstruction, $\theta_i$ denotes trainable parameters, $f_i$ contains the updated multi-level features, and $\mathcal{B}_i$ denotes the features stored in the feature bank. Features from preceding cascades are reused as $\mathcal{B}_1=\varnothing$ and $\mathcal{B}_i=\mathrm{Concat}(f_1,\ldots,f_{i-1})$ for $i>1$. Each cascade comprises FM feature refinement and a reconstruction network with data consistency.

\subsection{Foundation Model Feature Refinement}
The foundation model’s image encoder remains frozen throughout training. However, the distribution mismatch between initial FM features and native reconstruction features, together with artifacts in the input images, necessitates feature refinement before integration. Following common adaptation strategies~\cite{27}, we extract features from four layers across the encoder depth and refine them using lightweight adapters. Then the refined features are concatenated and fused by a convolutional neck module following MedSAM~\cite{6}.

To match the multi-level structure of the reconstruction network, the fused representation is projected into three feature branches corresponding to different reconstruction levels. A temporal-channel attention module~\cite{17}, adapted from channel attention by jointly modeling the temporal and channel dimensions, is applied to each branch to capture cine-specific spatio-temporal information. After that, features stored in $\mathcal{B}_i$ are fused with the corresponding feature branch through channel attention, further enriching the refined FM features used for integration.

\subsection{Cross-FiLM Feature Integration}
Naively injecting high-dimensional FM features into every cascade of an unrolled network via attention incurs substantial computational overhead. We therefore build upon FiLM~\cite{19}, which conditions features through lightweight channel-wise affine transformations, and propose Cross-FiLM to enable efficient bidirectional interaction between FM and native reconstruction features. Given FM features $F_{\mathrm{FM}}\in\mathbb{R}^{B\times T\times C\times H_{\mathrm{FM}}\times W_{\mathrm{FM}}}$ and cascade features $F_{\mathrm{CB}}\in\mathbb{R}^{B\times T\times C\times H_{\mathrm{CB}}\times W_{\mathrm{CB}}}$, where $B$ and $C$ denote batch size and channel number, global average pooling (GAP) produces $z_{\mathrm{FM}},z_{\mathrm{CB}}\in\mathbb{R}^{B\times T\times C}$. Linear projections generate modulation parameters $(w,b)$, and the bidirectional interaction is defined as
\begin{align}
F'_{\mathrm{FM}}
&=F_{\mathrm{FM}}
+w_{z_{\mathrm{CB}}}\odot F_{\mathrm{FM}}
+b_{z_{\mathrm{CB}}}, \\
F'_{\mathrm{CB}}
&=F_{\mathrm{CB}}
+w_{z_{\mathrm{FM}}}\odot F_{\mathrm{CB}}
+b_{z_{\mathrm{FM}}}.
\end{align}
Here, $F^{'}$ represents the updated feature maps, with $w$ and $b$ being the weight and bias embeddings produced from the feature embeddings via linear layers. Thus, native reconstruction features are conditioned on FM information, while FM features are simultaneously updated using reconstruction-specific representations. Then $F'_{\mathrm{CB}}$ continues through the reconstruction network, whereas $F'_{\mathrm{FM}}$ are stored in the feature bank and propagated to subsequent cascades.

\subsection{Training Objective}
Considering that both the k-space domain and image domain are critical for cine MRI reconstruction, we define the overall loss as $\mathcal{L} = \mathcal{L}_{\text{img}} +  \lambda_{1} \mathcal{L}_{\text{kspace}}$, where $\mathcal{L}_{\text{img}}$ and $\mathcal{L}_{\text{kspace}}$ denote the image-domain and k-space-domain loss terms, respectively. Here, $\mathcal{L}_{\text{kspace}}$ is defined as a mean squared error (MSE) loss, while $\mathcal{L}_{\text{img}}$ is formulated as a weighted combination of an $\ell_1$ loss, an MSE loss, and an SSIM loss: 
\begin{equation}
\begin{split}
    \mathcal{L}_{\mathrm{img}}
    ={}& \lambda_2 \|I_{\mathrm{rec}}-I_{\mathrm{gt}}\|_1
    + \lambda_3 \|I_{\mathrm{rec}}-I_{\mathrm{gt}}\|_2^2 \\
    &+ \lambda_4 \left(1-\mathrm{SSIM}
    (I_{\mathrm{rec}},I_{\mathrm{gt}})\right),
\end{split}
\end{equation}
where $I_{\mathrm{rec}}$ and $I_{\mathrm{gt}}$ denote the reconstructed and ground-truth images, respectively. For complex-valued outputs, SSIM is computed separately on the real and imaginary components and then averaged.

\section{Experiments and Results}
\subsection{Dataset}
We evaluated FM-Recon on the public CMRxRecon2023 dataset~\cite{14}, acquired on a 3T Siemens scanner and containing short-axis (SAX) and long-axis (LAX) cine MRI. Typical acquisition parameters include an in-plane resolution of $1.5\times1.5$ mm$^2$, FOV of $340\times340$ mm$^2$, slice thickness of 8 mm, and TR/TE of 3.6/1.6 ms. The dataset contains 120/60/120 subjects for training/validation/testing. We evaluated acceleration factors $R=4$, $8$, and $24$, with the $R=24$ mask generated following the same sampling scheme~\cite{14} to simulate a challenging high-acceleration setting.

\subsection{Implementation Details}
We used UNet-MR~\cite{24} with five cascades as the reconstruction backbone, owing to its competitive performance on the CMRxRecon2023 dataset at a moderate computational cost. Following~\cite{17}, five consecutive frames were used to reconstruct the center frame to decrease memory consumption without compromising reconstruction performance. To obtain FM inputs, zero-filled magnitude images were resized to $256\times256$ for SAM-based models and $224\times224$ for other foundation models, replicated to three channels, and normalized by their maximum intensity.

To assess the generalizability of FM-Recon across different FM representations, we evaluated six foundation models spanning three categories: SAM-based segmentation models (LiteMedSAM~\cite{6} and MCP-MedSAM~\cite{20}), vision–language models (CLIP~\cite{1} and PubMedCLIP~\cite{21}), and self-supervised vision models (DINOv2~\cite{3} and DINOv3~\cite{23}). Together, these models cover both medical and natural-image pretraining domains. FM features were integrated into all cascades for $R=4, 8$, whereas for $R=24$, were integrated from the third cascade onward, as determined by the following ablation study.

All models were implemented in PyTorch and trained on an NVIDIA RTX A6000 GPU (48 GB). For fair comparison, we used consistent hyper-parameters across experiments. We used AdamW with a learning rate of $2\times10^{-4}$, weight decay of 0.1, and batch size of 1. A cosine-annealing schedule with 10 warm-up epochs was applied over 80 epochs. The loss weights were set to $\lambda_1=0.25$, $\lambda_2=\lambda_3=0.5$, and $\lambda_4=1$. Reconstruction performance was evaluated using PSNR, SSIM, HaarPSI, and DISTS to provide a more comprehensive assessment from multiple perspectives, with background regions cropped before metric computation.

\subsection{Overall Comparison}
Table~\ref{tab1} compares FM-Recon using six foundation models with the UNet-MR baseline across different acceleration factors. Incorporating refined FM features consistently improves reconstruction performance, indicating that FMs can extract useful representations from artifact-corrupted inputs. The magnitude of improvement, however, varies across FMs. DINOv2 achieves the best overall performance at $R=4$ and $R=24$, while MCP-MedSAM performs best at $R=8$; in contrast, LiteMedSAM yields relatively minor gains. These results show that the effectiveness of FM features depends on both the choice of FM and the degree of undersampling, also indicating the importance of appropriate feature refinement to better exploit initial FM representations for MRI reconstruction.

\begin{table}[t]
\caption{Quantitative comparison of FM-Recon with different foundation models using UNet-MR. Metrics are computed on the cropped cardiac region. ``None'' denotes the baseline without FM features. Best results are in \textbf{bold}. $\dagger$: $p<0.05$ versus the baseline using a paired Wilcoxon signed-rank test with Bonferroni correction.}
\label{tab1}
\centering
\scriptsize
\setlength{\tabcolsep}{2pt}
\renewcommand{\arraystretch}{1.05}

\resizebox{\columnwidth}{!}{%
\begin{tabular}{c|l|c|c|c|c}
\hline
$R$ & FM 
& PSNR $\uparrow$ 
& SSIM $\uparrow$ 
& DISTS $\uparrow$ 
& HaarPSI $\uparrow$ \\
\hline

\multirow{7}{*}{$4\times$}
& None
& $40.99\pm2.86$
& $0.967\pm0.017$
& $0.939\pm0.017$
& $0.939\pm0.035$ \\

& LiteMedSAM
& $41.05\pm2.85^{\dagger}$
& $0.967\pm0.017^{\dagger}$
& $0.940\pm0.017^{\dagger}$
& $0.940\pm0.034^{\dagger}$ \\

& MCP-MedSAM
& $41.69\pm2.86^{\dagger}$
& $0.970\pm0.015^{\dagger}$
& $0.944\pm0.016^{\dagger}$
& $0.948\pm0.030^{\dagger}$ \\

& CLIP
& $41.45\pm2.84^{\dagger}$
& $0.969\pm0.015^{\dagger}$
& $0.943\pm0.016^{\dagger}$
& $0.946\pm0.031^{\dagger}$ \\

& PubMedCLIP
& $41.51\pm2.82^{\dagger}$
& $0.969\pm0.015^{\dagger}$
& $0.943\pm0.016^{\dagger}$
& $0.946\pm0.031^{\dagger}$ \\

& DINOv2
& $\mathbf{41.77\pm2.87}^{\dagger}$
& $\mathbf{0.971\pm0.015}^{\dagger}$
& $\mathbf{0.945\pm0.016}^{\dagger}$
& $\mathbf{0.949\pm0.031}^{\dagger}$ \\

& DINOv3
& $41.46\pm2.81^{\dagger}$
& $0.969\pm0.015^{\dagger}$
& $0.943\pm0.015^{\dagger}$
& $0.945\pm0.031^{\dagger}$ \\

\hline

\multirow{7}{*}{$8\times$}
& None
& $36.53\pm2.22$
& $0.933\pm0.022$
& $0.898\pm0.013$
& $0.870\pm0.043$ \\

& LiteMedSAM
& $36.71\pm2.21^{\dagger}$
& $0.935\pm0.022^{\dagger}$
& $0.901\pm0.014^{\dagger}$
& $0.873\pm0.042^{\dagger}$ \\

& MCP-MedSAM
& $\mathbf{37.10\pm2.23}^{\dagger}$
& $\mathbf{0.938\pm0.022}^{\dagger}$
& $\mathbf{0.904\pm0.014}^{\dagger}$
& $\mathbf{0.883\pm0.040}^{\dagger}$ \\

& CLIP
& $37.02\pm2.23^{\dagger}$
& $0.937\pm0.021^{\dagger}$
& $0.903\pm0.013^{\dagger}$
& $0.881\pm0.040^{\dagger}$ \\

& PubMedCLIP
& $36.82\pm2.23^{\dagger}$
& $0.936\pm0.022^{\dagger}$
& $0.903\pm0.013^{\dagger}$
& $0.877\pm0.041^{\dagger}$ \\

& DINOv2
& $37.07\pm2.25^{\dagger}$
& $\mathbf{0.938\pm0.021}^{\dagger}$
& $\mathbf{0.904\pm0.014}^{\dagger}$
& $0.882\pm0.040^{\dagger}$ \\

& DINOv3
& $37.06\pm2.24^{\dagger}$
& $\mathbf{0.938\pm0.021}^{\dagger}$
& $\mathbf{0.904\pm0.013}^{\dagger}$
& $0.882\pm0.040^{\dagger}$ \\

\hline

\multirow{7}{*}{$24\times$}
& None
& $32.35\pm1.85$
& $0.886\pm0.029$
& $0.851\pm0.014$
& $0.766\pm0.044$ \\

& LiteMedSAM
& $32.48\pm1.84^{\dagger}$
& $0.888\pm0.028^{\dagger}$
& $0.854\pm0.014^{\dagger}$
& $0.770\pm0.045^{\dagger}$ \\

& MCP-MedSAM
& $32.64\pm1.84^{\dagger}$
& $\mathbf{0.891\pm0.028}^{\dagger}$
& $0.857\pm0.014^{\dagger}$
& $0.775\pm0.043^{\dagger}$ \\

& CLIP
& $32.53\pm1.85^{\dagger}$
& $0.889\pm0.028^{\dagger}$
& $0.854\pm0.014^{\dagger}$
& $0.772\pm0.044^{\dagger}$ \\

& PubMedCLIP
& $32.58\pm1.83^{\dagger}$
& $0.890\pm0.028^{\dagger}$
& $0.855\pm0.014^{\dagger}$
& $0.774\pm0.043^{\dagger}$ \\

& DINOv2
& $\mathbf{32.73\pm1.84}^{\dagger}$
& $\mathbf{0.891\pm0.028}^{\dagger}$
& $\mathbf{0.858\pm0.013}^{\dagger}$
& $\mathbf{0.778\pm0.043}^{\dagger}$ \\

& DINOv3
& $32.60\pm1.84^{\dagger}$
& $0.890\pm0.028^{\dagger}$
& $0.854\pm0.014^{\dagger}$
& $0.775\pm0.040^{\dagger}$ \\

\hline
\end{tabular}%
}

\end{table}

\begin{figure*}[!tb]
\centering
\includegraphics[width=0.88\linewidth]{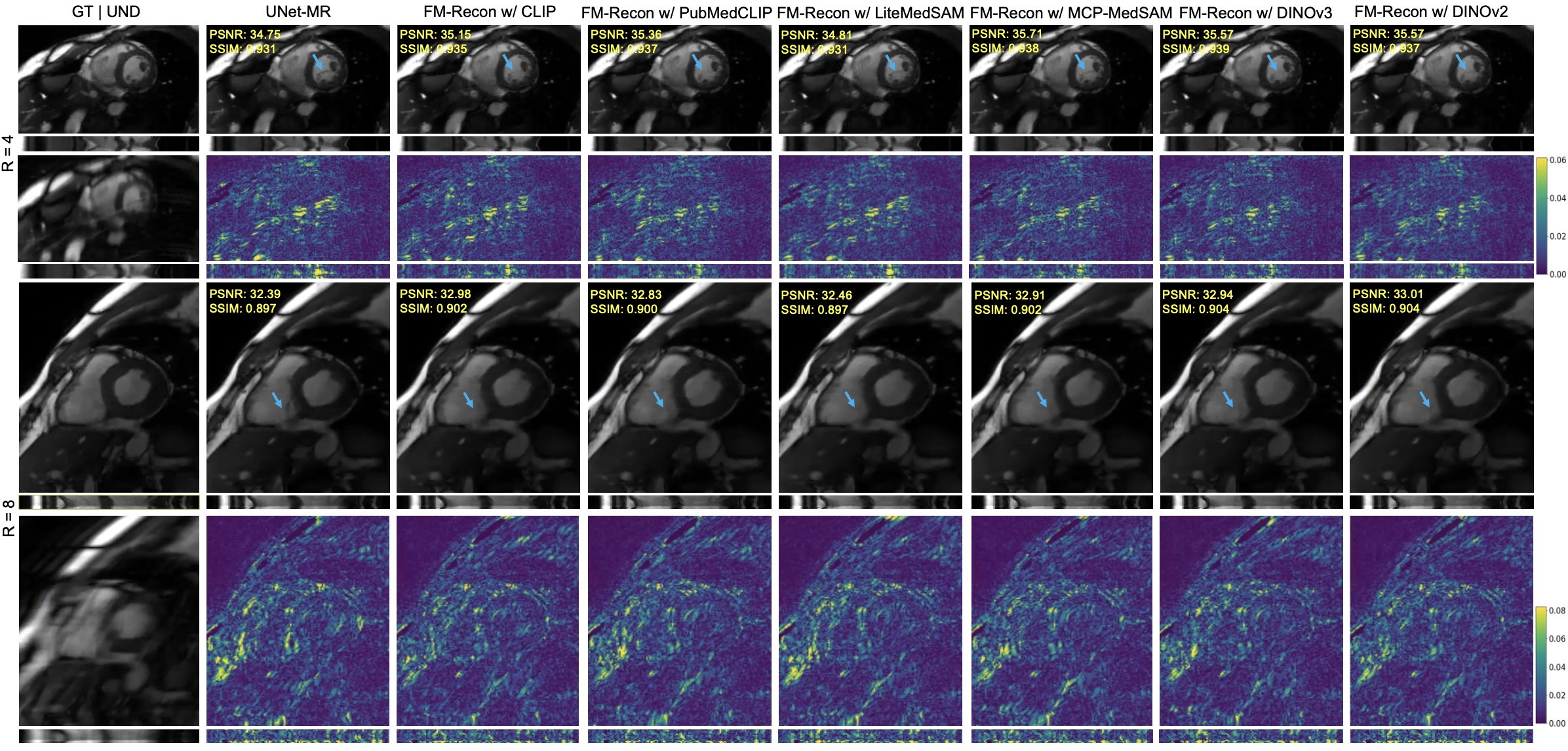}
\caption{Qualitative comparison of the UNet-MR baseline and FM-Recon with different foundation model backbones. Blue arrows highlight residual artifacts. The bottom row shows temporal profiles along the central spatial line, and the top-left values report metrics for the displayed region.}
\label{fig2}
\end{figure*}

Fig.~\ref{fig2} presents qualitative comparisons between the UNet-MR baseline and FM-Recon with different FM backbones. When the baseline exhibits residual artifacts (blue arrows), FM-Recon generally suppresses them to varying degrees depending on the FM used, while better preserving cardiac structures through the integration of refined FM features. The temporal profiles further suggest improved reconstruction consistency throughout the cine sequence.

\subsection{Ablation Study}
\begin{figure}[t]
\centering
\includegraphics[width=1\linewidth]{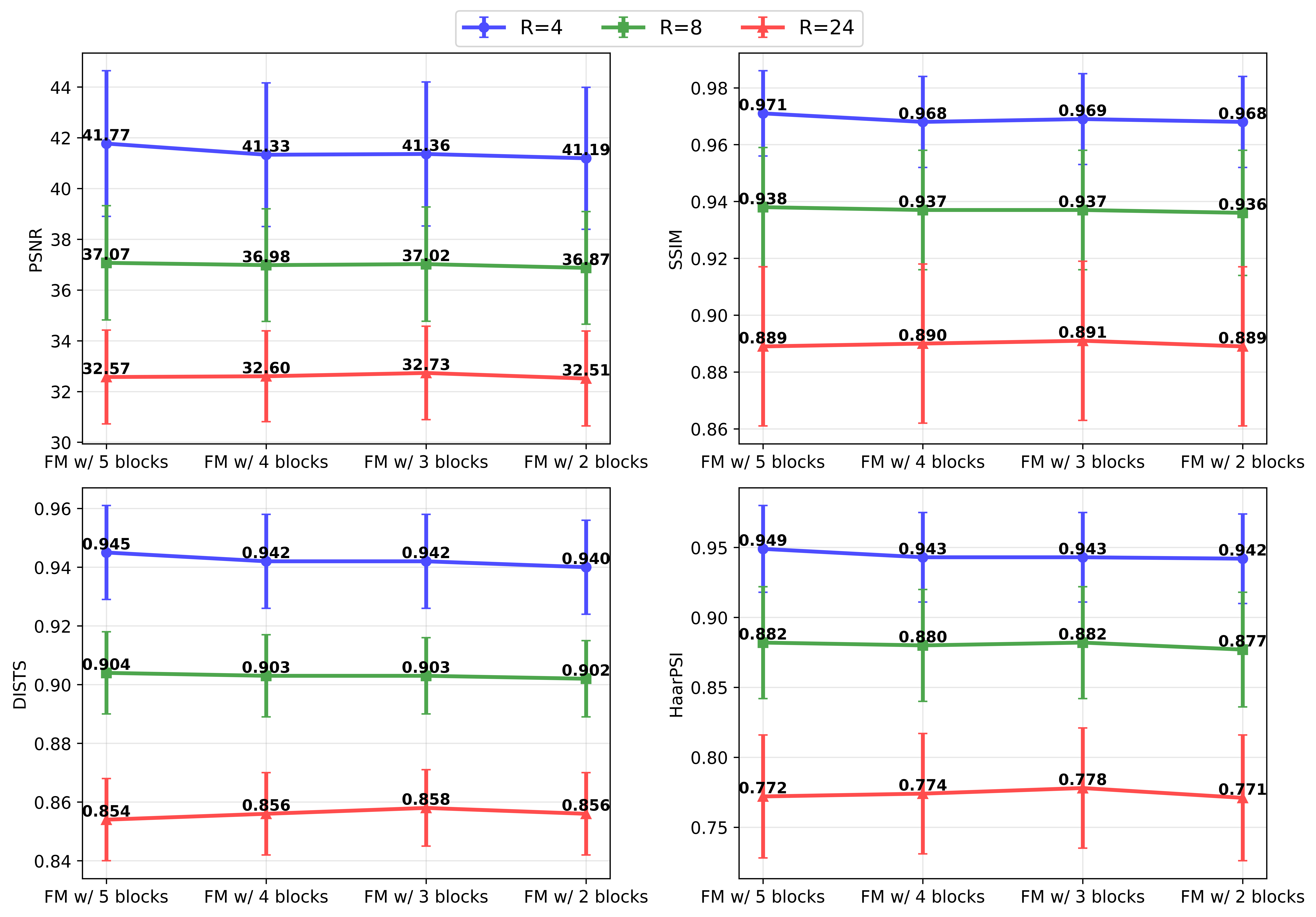}
\caption{Ablation study on the number of FM-integrated cascades using DINOv2 at different acceleration factors.}
\label{fig3}
\end{figure}

Although FMs can provide useful information from artifact-corrupted inputs, their effectiveness under varying artifact severity remains unclear. Since intermediate reconstructions become progressively cleaner along the unrolled network, we varied the number of FM-integrated cascades using DINOv2, one of the best-performing FMs, across acceleration factors. As shown in Fig.~\ref{fig3}, integrating FM features into all five cascades achieves the best performance at $R=4$. At $R=8$, integrating them into the last three to five cascades yields comparable results, suggesting diminishing benefits from FM features extracted at early reconstruction stages. This trend becomes more pronounced at $R=24$, starting integration from the third cascade performs best. These findings suggest that severe undersampling limits FM feature effectiveness and its impact cannot be fully suppressed through refinement, potentially degrading reconstruction performance.

\section{Discussion and Conclusion}
In this work, we explore the use of pretrained foundation model features for cardiac cine MRI reconstruction and propose FM-Recon, which refines and integrates these features into an unrolled reconstruction network. Our results show that useful FM features persist under moderate undersampling and can help improve reconstruction when appropriately refined and integrated. Furthermore, the effectiveness of FM-Recon depends on the chosen FM and artifact severity, with later integration proving more beneficial under severe undersampling. These findings highlight the potential of FM features for accelerated MRI reconstruction and offer insights into their effective integration.

\section{Compliance with Ethical Standards}
This study only used publicly available datasets~\cite{14}.

\section{ACKNOWLEDGMENTS}
This work is part of the ROBUST project (KICH3.LTP.20.006), partly funded by the Dutch Research Council (NWO), Philips Research, and the Dutch Ministry of Economic Affairs and Climate Policy (EZK) under the LTP KIC 2020--2023 program.
\bibliographystyle{IEEEbib}
\bibliography{strings,refs}

\end{document}